\documentclass[review,12pt]{elsarticle}
\usepackage{amssymb}
\usepackage{graphicx}
\usepackage{amsmath}

\usepackage{amssymb}
\usepackage{booktabs}
\usepackage{epsfig}
\usepackage{graphics}
\usepackage{epstopdf}
\RequirePackage{algorithmic}
\usepackage{booktabs}
\usepackage{color}
\usepackage{float}
\usepackage[capitalize]{cleveref}
\usepackage[ruled,linesnumbered]{algorithm2e}
\usepackage{subfig}
\usepackage{pdfpages}
\usepackage{tabularx}
\begin{document}

\begin{frontmatter}
\title{STHFL: Spatio-Temporal Heterogeneous Federated Learning}

\author[SEU,KLAB]{Shunxin~Guo}
\author[SEU,KLAB]{Hongsong~Wang}
% \author[SEU,KLAB]{Yuheng Jia\corref{cor}}
% \ead{yhjia@seu.edu.cn}
\author[SEU,KLAB]{Shuxia Lin}
\author[SEU,KLAB]{Xu Yang}
\author[SEU,KLAB]{Xin Geng\corref{cor}}
%\ead{xgeng@seu.edu.cn}
%\cortext[cor]{Corresponding authors}

%\affiliation[SEU]{organization={School of Computer Science and Engineering, Southeast University},%Department and Organization
%	city={Nanjing},
%	postcode={210096}, 
%	country={China}}
%
%\affiliation[KLAB]{organization={Key Laboratory of New Generation Artificial Intelligence Technology and Its Interdisciplinary Applications (Southeast University), Ministry of Education},%Department and Organization
%	country={China}}

\begin{abstract}
Federated learning is a new framework that protects data privacy and allows multiple devices to cooperate in training machine learning models. Previous studies have proposed multiple approaches to eliminate the challenges posed by non-iid data and inter-domain heterogeneity issues. However, they ignore the \textbf{spatio-temporal} heterogeneity formed by different data distributions of increasing task data in the intra-domain. Moreover, the global data is generally a long-tailed distribution rather than assuming the global data is balanced in practical applications. To tackle the \textbf{spatio-temporal} dilemma, we propose a novel setting named \textbf{Spatio-Temporal Heterogeneity} Federated Learning (STHFL). Specially, the Global-Local Dynamic Prototype (GLDP) framework is designed for STHFL. In GLDP, the model in each client contains personalized layers which can dynamically adapt to different data distributions. For long-tailed data distribution, global prototypes are served as complementary knowledge for the training on classes with few samples in clients without leaking privacy. As tasks increase in clients, the knowledge of local prototypes generated in previous tasks guides for training in the current task to solve catastrophic forgetting. Meanwhile, the global-local prototypes are updated through the moving average method after training local prototypes in clients. Finally, we evaluate the effectiveness of GLDP, which achieves remarkable results compared to state-of-the-art methods in STHFL scenarios.

\end{abstract}

\begin{keyword}
	Federated Learning \sep Spatio-Temporal Heterogeneity Data
%% keywords here, in the form: keyword \sep keyword

%% PACS codes here, in the form: \PACS code \sep code

%% MSC codes here, in the form: \MSC code \sep code
%% or \MSC[2008] code \sep code (2000 is the default)

\end{keyword}

\end{frontmatter}

%% \linenumbers

%% main text
\section{Introduction}
\label{sec:intro}

\begin{figure}[!htbp]
	\begin{center}
		\includegraphics[width=0.7\linewidth]{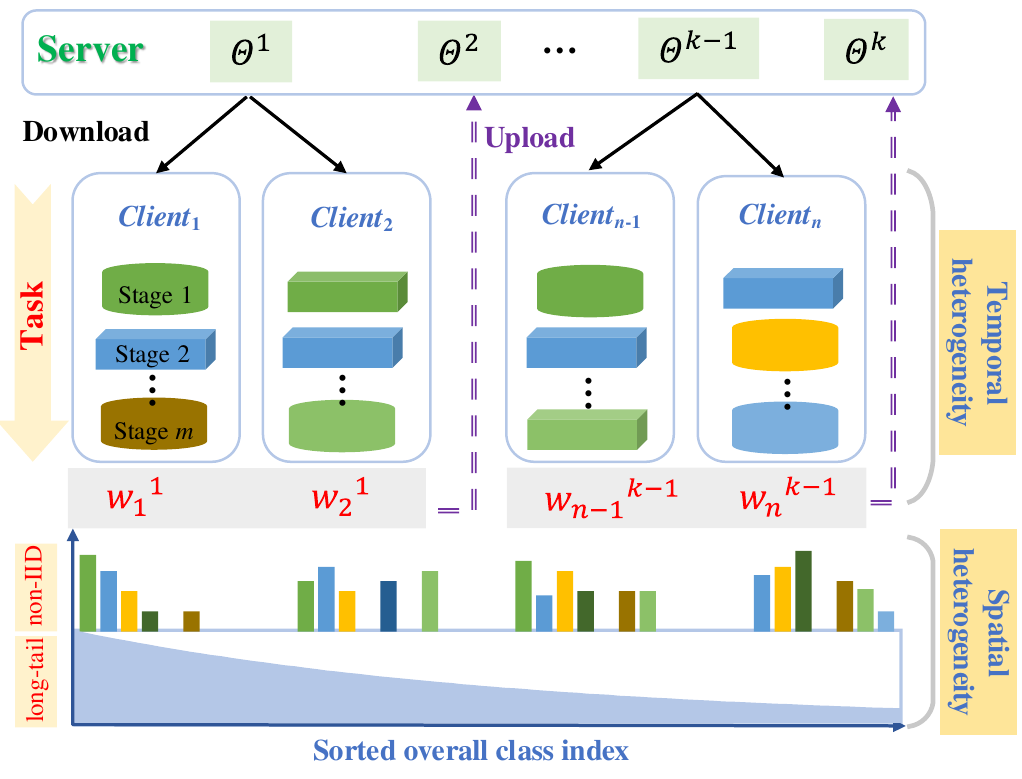}
	\end{center}
	\caption{
		The intuitive description of \emph{\textbf{spatio-temporal}} heterogeneous federated learning.
		(a) \emph{Spatial heterogeneity}: The local distribution across clients is non-iid,
		and the class distribution is long-tail.
		(b) \emph{Temporal heterogeneity}: The data inside the client is continuously increasing.
	}
	\label{fig:dlfl}
\end{figure}

Federated Learning (FL) enables different clients to collaboratively learn a global model, and it successfully addresses the challenge of data island while preserving client privacy~\cite{mcmahan2017communication}.
FL has attracted significant attention in various industrial applications~\cite{liu2020fedvision,zeng2021multi}, e.g., smart home~\cite{aivodji2019iotfla,li2020review}, autonomous vehicles~\cite{zeng2021multi}, and medical diagnosis~\cite{dong2020can}.
Due to the uniqueness of edge devices in complex real-world applications, the heterogeneity in federated learning has been a challenging research topic.
As shown in~\Cref{fig:dlfl}, there are two categories, one is \emph{spatial heterogeneity}, i.e., the data in distributed clients is non-identically distributed (\textbf{non-iid}) and the training data across clients is \textbf{long-tailed};
and another is \emph{temporal heterogeneity}, i.e., the data generated in each client is continuously increasing.
For example,
each medical institution builds a diagnostic model based on its own private medical records and stores the data locally.
Each institution has different disease distributions and different diseases make up a long-tailed distribution as a whole, some diseases are common and others are rare (i.e., \emph{spatial heterogeneity}).
Moreover, the number of cases in each facility increased dynamically over time
(i.e., \emph{temporal heterogeneity}).

To address the non-iid~\cite{zhao2018federated} problem, many methods have been proposed,
including correcting label distribution deviation from the statistical perspective~\cite{zhang2022federated,chen2020fedbe}, constructing the personalized model based on representation learning\cite{finn2017model, li2019fedmd, wang2019federated,liang2020think, collins2021exploiting, li2021ditto}, and distilling global knowledge to local model~\cite{afonin2021towards,zhang2022fine}.
These methods solve the issues of local model drift and global model performance degradation~\cite{hsu2019measuring,khaled2020tighter,lee2021preservation} caused by non-iid.
Along with its pilot progress, researchers found that the cross-client data has a global long-tail distribution in the real FL scenario.
If the training data across clients is both long-tailed and non-iid,
the joint problem becomes complex because each client may hold classes with fewer samples.
The biased local models perform poorly on different classes and directly affect the aggregated global model.
Existing FL methods usually cited traditional unbalanced learning methods~\cite{cao2019learning, dredze2010multi, dredze2008online, yang2021federated} or employed the interactive advantages of the FL framework itself~\cite{shang2022federated,2022Federated,afonin2021towards}.
Although most FL methods that focus on the \emph{spatial heterogeneity} challenge indeed solve the relevant problems,
they typically assume that the data of the whole FL framework is fixed and known in static application scenarios, and they ignore that the local clients receive new data in an online manner (\emph{temporal heterogeneity}).
%However,these FL methods are assuming that the data of the whole FL framework is fixed and known in static application scenarios, they ignore that the local clients receiving new data in an online manner (\emph{temporal heterogeneity}).

Therefore, we consider the more challenging setting which is \emph{\textbf{spatio-temporal}} heterogeneous federated learning in this paper.
Among them, the data distributions across clients are different and the dataset globally is unbalanced (\emph{spatial heterogeneity}).
Moreover, each client contains multiple stage tasks, in which data distributions between the current stage and the previous stage tasks are different (\emph{temporal heterogeneity}).
The \emph{spatial heterogeneity} may lead to inter-domain concept shift and biased global-local models on global long-tail distribution, and the \emph{temporal heterogeneity} may lead to the intra-domain concept shift and catastrophic forgetting~\cite{kirkpatrick2017overcoming,rebuffi2017icarl} of local models.
In addition, the temporal heterogeneity exacerbates the non-iid problem in spatial heterogeneity.

To tackle these challenges in Spatio-Temporal Heterogeneous FL (STHFL), we propose a novel Global-Local Dynamic Prototype model (GLDP) in this paper, which effectively ensures that diverse client participants can train the global-local unbiased models with strong generalization ability while reducing local catastrophic forgetting.
To solve the inter-domain shift problem of \emph{spatial heterogeneity}, the participating clients train a common underlying representation and their respective personalized representations.
The generalized common representation can improve the generalization of the global model, while the personalized representation for adapting different clients.
The global prototype can be used as supplementary knowledge of classes to construct the unbiased local model which can reduce the influence of long-tailed distribution in \emph{spatial heterogeneity}.
As for \emph{temporal heterogeneity}, we establish local prototype relation distilling to address the catastrophic forgetting problem caused by the intra-domain drift and dynamically update global-local prototypes using moving average method to balance old and new knowledge.
We demonstrate the effectiveness of proposed GLDP method, and it outperforms other personalized FL methods in spatio-temporal heterogeneous FL setting.
The main contributions are as follows:
\begin{itemize}
	\item We design a new Spatio-Temporal Heterogeneity FL setting formed by the intersection of \emph{temporal heterogeneity} within each client and \emph{spatial heterogeneity} cross clients.
	\item We propose a novel framework named GLDP, in which the personalized model combined with global and local dynamic prototypes address the problems in STHFL.
	\item We introduce the feasibility of prototype knowledge to protect data privacy and the indispensability of effective communication.
\end{itemize}

%-------------------------------------------------------------------------
\section{Related work}
\textbf{Federated with heterogeneous data.}
The FedAvg~\cite{mcmahan2017communication} emphasizes that client training data is non-iid and imbalanced data samples among clients are two classic problems that are prevalent in the real world.
\par
\textbf{For non-iid,}~\cite{2020Personalized} studies a personalized variant of federated learning.
It provides each client with a more personalized model through local training based on the meta-model.
The heterogeneity of the model indirectly caused by the heterogeneity of client data, ~\cite{2020Ensemble} promotes the use of federated distillation for model fusion.
To avoid the existence of data on the server side, FedGEN~\cite{2021Data} proposed a server-side knowledge distillation method without data for heterogeneous federated learning.
The core idea is to extract knowledge about the global view of the data distribution to adjust local model training.
In addition, researchers have proposed some schemes based on cluster sampling, mainly according to the nature of the learning task or the similarity of the data samples~\cite{caldarola2021cluster,taik2022clustered,sattler2020clustered}.
These methods focus on selecting representative clients and reducing the variance of randomly selected customer aggregation weights~\cite{2020An,2021Clustered,2020Briggs}.
Although the abovementioned methods address non-iid data to a certain extent~\cite{bengio2013representation,lecun2015deep,collins2021exploiting}, they perform poorly on classes with few samples because the lack of consideration of the global long-tailed distribution.
\par
\textbf{For non-iid and long-tail distribution,} GRP-FED~\cite{2021GRP} designed an adaptive aggregation parameter to adjust the aggregation ratio to ensure fairness for different clients.
Shang et al.~\cite{shang2022federated} presented methods to retrain heterogeneous and long-tailed data by a classifier with joint features.
Concurrently, they also proposed a new distillation method with logit adjustment and calibration gating network~\cite{2022Federated}.
Ratio loss~\cite{afonin2021towards} estimates the global class distribution based on balanced auxiliary data on the server, in order to make local optimization equivalent to a balanced state.
This method requires necessary relevant data on the server side to train the model.
However, collecting data on the server side may reduce data security and be impractical.

\textbf{Continual learning in Federated learning.}
The purpose of continual learning~\cite{shmelkov2017incremental, li2017learning, yang2021flop,pourkeshavarzi2021looking} is to continuously learn new classes while alleviating the forgetting problem of old classes.
Continual learning tasks also exist in federated learning frameworks, where data is generated dynamically due to actual applications.
The solution to this problem is usually to store data of old categories on the client and then interact with the server to get a global model~\cite{yang2021flop}.
Recent research~\cite{yoon2021federated} is that each client selects the required knowledge from other clients through a weighted combination of task-specific parameters.
GLFC~\cite{dong2022federated} utilizes a loss function to balance forgetting of old classes and extract consistent inter-class relationships across tasks.
{\color{black}Similarly, CFeD~\cite{macontinual} requires both the client and the server to independently have the unlabeled proxy dataset to reduce forgetting.
	Based on realistic client enrollment
	scenarios, ~\cite{park2021tackling} proposes to use deep variational embeddings to preserve the privacy of customer data while~\cite{hendryx2021federated} utilizes prototypical networks to enable per-customer class incremental learning.
}
These methods are based on domain-incremental and class-incremental learning research, and most require additional memory to save old class data for solving the catastrophic forgetting problem.
We differ from them in the \emph{temporal heterogeneity} formed by task increments within each client, and also consider the complexity of spatial heterogeneity among clients.

%-------------------------------------------------------------------------
\section{Problem Definition}

Given the input space $\mathcal{X}$ and the output space $\mathcal{Y}$,
let $P$ be the distribution of $\mathcal{X} \times \mathcal{Y}$.
For local clients $\{F_i\}_{i = 1}^N$
in Spatio-Temporal Heterogeneous Federated Learning (STHFL),
we have an associated dataset
$\mathcal{D}_{i}$ which contains $M$ stage tasks.
And dataset in each stage task is 
$\mathcal{D}_i^{(m)} = \left\{\left(X_{i}^{(m)}, Y_{i}^{(m)}\right) \mid X_{i}^{(m)} \in \mathbb{R}^{T_{i}^{(m)} \times U}, Y_{i}^{(m)} \in \mathbb{R}^{T_{i}^{(m)} \times z_i^{(m)}}\right\}$,
where $X_{i}^{(m)}$ and $Y_{i}^{(m)}$ denote the private input matrix and output matrix respectively, $T_{i}^{(m)}$ is the number of private data in the $m$-th stage of client $i$,
$U$ represents the input space, 
and $z_i^{(m)}$ is the number of classes in the $m$-th stage. 
What's more,
we use $P_{(X_{i}^{(m)},Y_{i}^{(m)})}$ as the data distribution in $m$-th stage task for $i$-th client.
We also denote the central server as $F_G$.
The \emph{spatial heterogeneity} of inter-client and \emph{temporal heterogeneity} of each client are defined as following:
\begin{itemize}
	\item \emph{Spatial heterogeneity}:
	\begin{itemize}
		\item $P_{(X_{i}^{(m)},Y_{i}^{(m)})} \neq P_{(X_{j}^{(m)},Y_{j}^{(m)})}$.
		The data distributions in $m$-th stage task for all client are different,
		which may lead to inter-domain concept shift.
		\item The data for each class is unbalanced in any stage tasks of clients.
		Moreover, from a global perspective,
		the data distribution is long-tailed.
	\end{itemize}
	%  \\(1) $\mathcal{P}_i(X|\mathcal{Y}) \neq \mathcal{P}_j(X|\mathcal{Y})$. There exists inter-domain concept shift, the conditional distribution  $\mathcal{P}(X|\mathcal{Y})$ of data across local clients even if they follow the ${P}(\mathcal{Y})$.\\
	% (2) $\mathcal{P}_i(\mathcal{Y}) \neq \mathcal{P}_j(\mathcal{Y})$. There exists inter-domain label distribution skew with different clients.
	\item \emph{Temporal heterogeneity}: 
	\begin{itemize}
		\item$P_{(X_{i}^{(m)},Y_{i}^{(m)})} \neq P_{(X_{i}^{(m-1)},Y_{i}^{(m-1)})}$.
		The data distribution of each stage task is different for all clients,
		which may cause intra-domain concept shifts, especially knowledge shifts in each class.
		\item$\exists y_{i}^{(m),r} \notin \bigcup_{j=1}^{(m-1)} Y_{i}^{(j)} \text{ for }  y_{i}^{(m),r} \in Y_{i}^{(m)}$.
		The classes in the current stage task may not be seen in previous tasks, the parameters of the model will be rewritten after training on the current task which causes catastrophic forgetting.
	\end{itemize}
\end{itemize}
In addition,
the central global server can access the prototype knowledge of all clients.
The aim of STHFL is to make the models in clients have strong generalization for intra-domain concept shift and 
prevent catastrophic forgetting.
Meanwhile,
the global model can help clients learn collaboratively to tackle non-iid and long-tailed data distribution.

%-------------------------------------------------------------------------
\section{The Proposed GLDP Model}
\begin{figure*}[!ht]
	\begin{center}
		\includegraphics[width=1\linewidth]{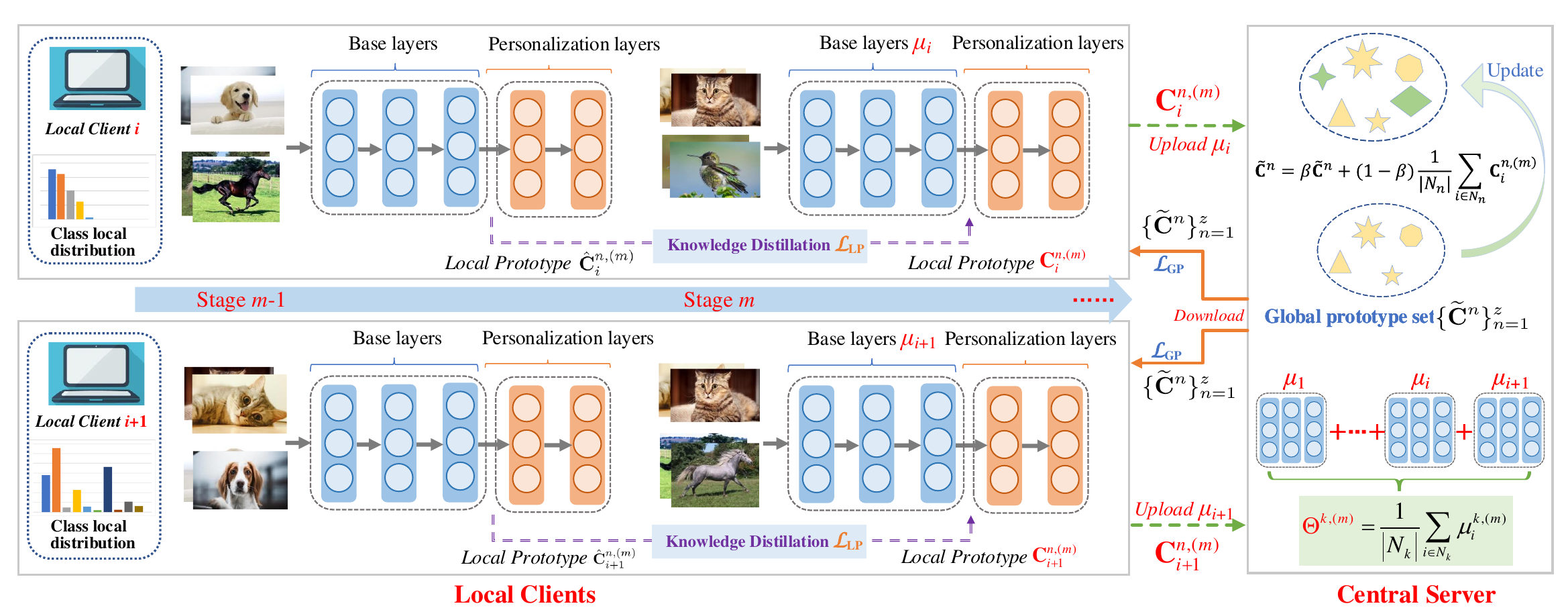}
		\vspace{-0.5cm}
	\end{center}
	\caption{Illustration of GLDP.
		Simplified schematization of our method that solves spatio-temporal heterogeneous federated learning problem via \emph{Federated representation learning} and \emph{Federated prototype learning}.
		(1) \emph{Federated represention learning.}
		The local client trains the personalization model to get the shared representation layer ($\mu$) and the personalization layer to fit the non-iid distribution.
		(2) \emph{Federated prototype learning.} $\mathcal{L}_{\mathrm{LP}}$: Guide the learning of post-stage $\mathbf{C}_{i}^{n,(m)}$ based on preserved local prototype knowledge $\hat{\mathbf{C}}_{i}^{n,(m)}$ to reduce local model catastrophic forgetting inherent to temporal heterogeneity.
		$\mathcal{L}_{\mathrm{GP}}$: By making the local prototype approximate the global prototype to learn global knowledge, reduce the global-local model bias generated by long-tail.
		Compute the global model parameter $\Theta ^{k,(m)}$ and update the global prototype set $ \{\widetilde{\mathbf{C}}^{n}\}^z_{n=1}$ based on uploaded $\{\mathbf{C}^{n,(m)}_{i}\}_{i=1}^{N_k}$ on the central server.}
	\label{fig:short}
	\vspace{-0.5cm}
\end{figure*}

The overview of our model is depicted in~\Cref{fig:short}.
To address the problem of Spatio-Temporal Heterogeneous Federated Learning (STHFL), our model mitigates non-iid via representation learning (\Cref{person}) and using local pre-stage and post-stage prototype relations to solve \emph{temporal heterogeneity}, while using global-local prototype relations to alleviate long-tail distribution (\Cref{proto}).

\subsection{Federated Representation Learning}\label{person}
In the basic FL setting, the participating clients are independent and identically distributed and their private data is fixed. Therefore, all local clients share the same network structure with the same parameters, and the basic objective function is:
\begin{equation}
	\begin{split}
		\label{eq:FEDAVG}
		\underset{\omega}{\arg \min }\frac{1}{N} \sum_{i=1}^{N}  \mathcal{L}_{CE}\left(\mathcal{F}_\omega\left( X_i\right), Y_i\right),
	\end{split}
\end{equation}
where $\mathcal{L}_{CE}$ represents the Cross-Entropy loss for supervised learning, and $\mathcal{F}(\cdot)$ expresses the shared model among clients.

However, in the more challenging STHFL setting, each client's local model does not share the same structure to deal with the spatio-temporal heterogeneity. Thus we use $\mathcal{F}_{\omega_i}(\cdot)$ to denote the $i$-th client model to strengthen such difference. Therefore, the classification loss of $m$-stage on the $i$ client is rewritten as:
\begin{equation}
	\begin{split}
		\label{eq:FEDAVG}
		\mathcal{L}^{i,{(m)}}_{CE} = \mathcal{L}_{CE} (\mathcal{F}_{\omega_i^{(m)}}(X_i^{(m)}), Y_i^{(m)}),
	\end{split}
\end{equation}
where $\omega_i^{(m)}$ is the corresponding local model parameter that can interact with the server. 

%and $\mathcal{D}^{(m)}$ is the set of all samples of participating clients at the $m$-th stage. %($D^{(m)} = \left|T^{(m)}\right| $)%

For the $i$-th client model $\mathcal{F}_{\omega_i}(\cdot)$, it contains two parts which are the representation layer $f_{\mu_i}\left(\cdot\right)$ and the classification layer $g_{\nu_i}\left(\cdot\right)$. 
For $f_{\mu_i}\left(\cdot\right)$, it works to transform the inputs from diverse clients into the corresponding embedding spaces, which will contribute to a global shared representation space after collaborative learning.  For $g_{\nu_i}\left(\cdot\right)$, it classifies the inputs of each client to the corresponding specified label space.
% (1) The former transforms the inputs from diverse clients into the shared embedding space.
% % refers to transforming the input from the original feature space to the embedding space, enabling tasks of different distributions to share the common representation.
% In general, the embedding function of the $i$-th client is $f_{\mu_i}\left(\cdot\right)$ parameterized by $\mu_{i}$.
% %and denotes $\mathcal{H}_{i}=f_{i}\left(\mu_{i} ; \mathbf{x}\right)$ as the embedding of $\mathbf{x}$.
% (2) The latter refers to the decision-making layer that can make classification decisions.
% %If a supervised learning task implements the classification of samples,
% The classification of $\mathbf{x}$ can be done by the function $g_{\nu_i}\left(\cdot\right)$ parameterized by $\nu_{i}$.
Therefore, the network of each client $\mathcal{F}_{\omega_i}(\cdot)$ can be re-written as:
$\mathcal{F}_{\mu_{i}, \nu_{i}}= f_{\mu_i}\circ g_{\nu_{i}}$, where $\omega_i$ is an abbreviation for $\left(\mu_{i},\nu_{i}\right)$.

As with most federated learning, GLDP also contains multiple global rounds.
In the $k$-th global round, we locally update $\mu$ and $\nu$ with different times in the local model to accommodate clients with non-iid data.
%We train the local model to obtain the personalization layer of each client and a generalizable representation across clients.
The base representation layers $\mu_{i}^{k}$ of client $i$ update the local gradient:
\begin{equation}
	\begin{split}
		\label{eq:share}
		\mu_{i}^{k, t+1}=\operatorname{grad}\left(\mathcal{F}_{\mu_{i}^{k, t}, \nu_{i}}, \mu_{i}^{k, t}, \alpha\right),
	\end{split}
\end{equation}
where $\alpha$ represents the step size, $t$ is the number of rounds of local model updating, and $\operatorname{grad}(\mathcal{F}_i, \mu_i, \alpha)$ is the general function of gradient updating for the variable $\mu_i$ in local function model $\mathcal{F}_i$.
After each participating client completes the local update, iteratively communicates with the server to obtain the shared presentation layer.
It contains the relevant information of multiple domains, which can reduce the inter-domain concept shift of \emph{spatial heterogeneity}.

The personalization layer of each client contains its own private special class information, which can adapt to their respective class distribution and thus mitigate inter-domain label distribution skew of non-iid.
The updated formula is defined as:
\begin{equation}
	\begin{split}
		\label{eq:per}
		\nu_{i}^{k,s+1}=\operatorname{grad}\left(\mathcal{F}_{\mu_{i}^{k,s}, \nu_{i}}, \nu_{i}^{k,s}, \alpha\right),
	\end{split}
\end{equation}
where $s$ is the number of rounds of local update of the personalization layer, generally $t < s$.

Furthermore, we only transfer the shared presentation layer $\mu$ to the server-side aggregation due to the nature of FL protecting private data.
At the $m$-th stage of $k$-th global round, the aggregating the global model $\Theta ^{k,(m)}$ formula on the \textbf{server-side} is:
\begin{equation}
	\begin{split}
		\label{eq:agg}
		\Theta ^{k,(m)}=\frac{1}{\left |N_k  \right | } \sum_{i \in N_k} \mu_{i}^{k,(m)},
	\end{split}
\end{equation}
where $N_k$ is the set of selected clients and $\mu_{i}^{k,(m)}$ is the shared layer parameters of $m$-th stage.

\subsection{Federated Prototype Leaning}\label{proto}
To address the \emph{temporal heterogeneity} and long-tail problem, we bring ideas from the prototype theory~\cite{snell2017prototypical,tan2022federated,michieli2021prototype, mu2021fedproc} to propose two novel training objectives, which are local ($\mathcal{L}_{\mathrm{LP}}$) and global-local prototype relations ($\mathcal{L}_{\mathrm{GP}}$). In the GLDP framework, we learn the prototype-related knowledge from the data of all the clients and then pass this prototype to the central server. Specifically, for each label, we have a prototype which is calculated by averaging the embedding vectors of the same label in different clients. Since the average is an irreversible process~\cite{tan2022fedproto} that it is impossible to reconstruct the original data from the prototype when the attacker does not have access to the local model data, the pass of the prototype will not invalidate the privacy protection. 

Formally, let $\mathbf{C}^{n}$ be the $n$-th class prototype and the input class space is $z$.
At the $m$-stage, the definition of the prototype for client $i$ is as follows:
\begin{equation}
	\begin{split}
		\label{eq:lp1}
		\mathbf{C}^{n,(m)}_{i}=\frac{1}{\left|\mathcal{D}^{(m)}_{i,n}\right|} \sum_{\left(\mathbf{x}, y\right) \in \mathcal{D}^{(m)}_{i, n}} f_{\mu_{i}^{(m)}}\left(\mathbf{x}\right),
	\end{split}
\end{equation}
where $\mathcal{D}^{(m)}_{i,n}$ represents the sample set of class $n$ on the $i$-th client {\color{black}
	and $f_{\mu_{i}^{(m)}}\left(\mathbf{x}\right)$ denotes the representation layer of sample $\mathbf{x}$}.
Then, the local prototype set of this stage on the $i$-th client is defined as $\{\hat{\mathbf{C}}_i^{n,(m)}\}^{z_i^{(m)}}_{n=1}$.
{\color{black}
	The global prototype set is gradually updated according to the collaborative learning among clients to obtain$\{\widetilde{\mathbf{C}}^{n}\}^z_{n=1}$.
}
%nd gradually updated according to the local prototype to obtain $\{\widetilde{\mathbf{C}}^{n}\}^z_{n=1}$.

\noindent\textbf{Local prototype relation.}
For each client, the tasks keep varying data distributions, this is the so-called \emph{temporal heterogeneity}, which will cause the intra-domain label distribution skew problem that leads to catastrophic forgetting. To against it, we design a local prototype relation loss $\mathcal{L}_{\mathrm{LP}}$ to guide the prototype at the new stage by the one at the previous stage. 
If the local prototype of class $n$ exists at the $i$-th client, the loss at the $m$-th stage is defined as:
\begin{equation}
	\begin{split}
		\label{eq:KL1}
		\mathcal{L}^{i,{(m)}}_{\mathrm{LP}}=\mathrm{KL}\left(\hat{\mathbf{C}}_{i}^{n,(m)}\| \mathbf{C}_{i}^{n,(m)}\right),
	\end{split}
\end{equation}
where $\mathrm{KL}(\cdot \| \cdot)$ is Kullback-Leibler divergence and $\hat{\mathbf{C}}_{i}^{n,(m)}$ represents the locally old class $n$ prototype of previous stages.
The $\mathbf{C}_{i}^{n,(m)}$ can obtain the corresponding prediction probability ${p}_{i}^{n,(m)}$ through the training of the class data samples $\mathbf{x}_{i}^{n,(m)}$ and the local model $\omega_{i}^{(m)}$, which is
\begin{equation}
	\begin{split}
		\label{eq:KL2}
		\mathcal{L}^{i,{(m)}}_{\mathrm{LP}} =\mathrm{KL}\left(\hat{\mathbf{C}}_{i}^{n,(m)}\| p_{i}^{n,(m)}\left(\mathbf{x}_{i}^{n,(m)}, \omega _{i}^{(m)}\right)
		\right).
	\end{split}
\end{equation}
This regularization term updates the local model at the new stage and thus minimizes the intra-domain concept drift.

Since our setting is task increment within the client and does not save data samples from previous tasks, the $n$-th local prototype is updated to retain the learned knowledge of the old class new samples according to the moving average~\cite{gu2010detrending} as follows:
\begin{equation}
	\begin{split}
		\label{eq:move}
		\hat{\mathbf{C}}_{i}^{n,(m+1)} = \beta \hat{\mathbf{C}}_{i}^{n,(m)} + (1 -\beta) {\mathbf{C}}_{i}^{n,(m)},
	\end{split}
\end{equation}
{\color{black}where $\beta$ is a moving average hyperparameter of the local prototype.}
Finally, we update the local prototype set $\{\hat{\mathbf{C}}_i^{n,(m)}\}^{z_i^{(m)}}_{n=1}$, where the prototype of the new class is added directly.

\noindent\textbf{Global-local prototype relation.}
The long-tailed distribution of classes will directly affect the generalization of the local model intra-domain and indirectly lead to the bias of the global model.
To alleviate such challenge, we construct a global-local prototype relation loss to make the local prototype approximate to the global prototype. In this way, the biased local model caused by the class with few samples is reduced. Such loss is:
% in order to reduce the biased local model caused by the class with few samples, as follows:
\begin{equation}
	\begin{split}
		\label{eq:GP}
		\mathcal{L}^{i,{(m)}}_{\mathrm{GP}} = \sum_{n\in{z_i^{(m)}}} \frac{T_i^{n,(m)}}{T_i^{(m)}} \mathcal{L}_\mathrm{MSE}\left(\mathbf{C}^{n,(m)}_{i}, \widetilde{\mathbf{C}}^{n}\right),
	\end{split}
\end{equation}
where $T_i^{n,(m)}$ is the number of instances belonging to class $n$ over $i$-th client and $\mathbf{C}^{n,(m)}_{i}$ denotes the local prototype of the $n$-th label.
%, and ${\mathrm{MSE}}(\cdot,\cdot)$ represents the Mean Squared Error loss.
Besides, it can reduce the catastrophic forgetting of the local model by learning new classes of relevant information based on collaboration among clients.
Then, since no data samples exist on the server, the global prototype of class $n$ is updated on \textbf{server-side} similar as \cref{eq:move}:
\begin{equation}
	\begin{split}
		\label{eq:GP1}
		\widetilde{\mathbf{C}}^{n}= \beta\widetilde{\mathbf{C}}^{n} + (1-\beta) \frac{1}{\left|N_{n}\right|} \sum_{i \in N_{n}} \mathbf{C}^{n,(m)}_i,
	\end{split}
\end{equation}
where $N_{n}$ is the set of clients that have class $n$.
Finally, we update the global prototype set $\{\widetilde{\mathbf{C}}^{n}\}^{z}_{n=1}$ and directly add the new class prototype.

In summary, the optimization objective for the $m$-th stage of $i$-th local client is
\begin{equation}
	\begin{split}
		\label{eq:taotal}
		\mathcal{L}^{i,{(m)}}_{\mathrm{Total}}  = \mathcal{L}^{i,{(m)}}_{CE}+ \lambda \cdot \mathcal{L}^{i,{(m)}}_{\mathrm{LP}} + (1-\lambda) \cdot \mathcal{L}^{i,{(m)}}_{\mathrm{GP}},
	\end{split}
\end{equation}
{\color{black}
	Where $\lambda$ is a hyperparameter representing the ratio of the two loss functions of $\mathcal{L}_\mathrm{LP}$ and $\mathcal{L}_\mathrm{GP}$.} We update the local model via optimizing~\cref{eq:taotal}, and then aggregate parameter $\mu_i^{(m)}$ and local prototype $\mathbf{C}_{i}^{n,(m)}$ in central server $F_G$ to obtain the global model and global prototype set for the next stage.

At the test stage, we use the distance between the test sample's embedding representation vector $f_{\mu_i}\left(\mathbf{x}^{test}\right)$ and the prototype to obtain the predicted label $\hat{y}$ of sample $\mathbf{x}^{test}$ as follows:
\begin{equation}
	\begin{split}
		\label{eq:test}
		\hat{y}=\underset{n}{\arg \min }\left\| f_{\mu_i}\left(\mathbf{x}^{test}\right)-\mathbf{C}_i^{n}\right\|_{2},
	\end{split}
\end{equation}
where $\mathbf{C}_i^{n}$ is the $n$-th local prototype of $i$-th client, and can be replaced with the 
$\mathbf{C}^{n}$ of the $n$-th global prototype on testing.
\subsection{The Optimization Process of GLDP}

\begin{algorithm}[!ht]\footnotesize
	\caption{GLDP}\label{algorithm: AllModel}
	\textbf{Input}: At the $k$-th global round, $F_G$ randomly selected the $N_k$ local clients; Each client includes $M$ tasks data $F_i = \{F^{(1)}_i, \ldots, F^{(m)}_i,\ldots, F_i^{(M)}\}$; $F_G$ sends the latest global model parameter $\Theta ^{k-1}$ and global prototype set $\{\widetilde{\mathbf{C}}^{n}\}^z_{n=1}$ to $N_k$ clients;\\
	\begin{algorithmic}[1]
		\STATE $\Theta^{k,(m)} \gets  \Theta^{k-1}$ and $\{\widetilde{\mathbf{C}}^{n,k}\}^z_{n=1} \gets \{\widetilde{\mathbf{C}}^{n}\}^z_{n=1}$
		\FOR{$m$ = 1 : $M$ }
		\STATE \textbf{Central Server:}
		\STATE Local Clients Update($\{\widetilde{\mathbf{C}}^{n,k}\}^z_{n=1}$, $\Theta^{k,(m)}$);
		\STATE Update global model parameters $\Theta^{k,(m)}$ via~\cref{eq:agg};
		\STATE Update global prototype $\widetilde{\mathbf{C}}^{n}$ by~\cref{eq:GP1};
		\STATE Update global prototype set $\{\widetilde{\mathbf{C}}^{n}\}^z_{n=1}$;\\
		\STATE $\Theta^{k,(m+1)} \gets \Theta^{k,(m)}$;
		\STATE \textbf{Local Clients Update($\{\widetilde{\mathbf{C}}^{n,k}\}^z_{n=1}$, $\Theta^{k,(m)}$):}\\
		\FOR{each client $n$ of $N_k$ \textbf{in parallel}}
		\FOR{each local epoch $r$ = 1 : $t+s$}
		\STATE Update local model base layers $\mu_i^{(m)}$ by ~\cref{eq:share} \textbf{if} $r = 1 : t$;\\
		\STATE Update local model personalized layers $\nu_i^{(m)}$ by ~\cref{eq:per} \textbf{if} $r = t+1 : t+s$;
		\STATE Obtain local model parameters $\omega_i^{(m)}$ and local prototype $\mathbf{C}_i^{n,(m)}$ by optimizing ~\cref{eq:taotal};
		\STATE Update local prototype set $\{\hat{\mathbf{C}}_i^{n,(m)}\}^{z^{(m)}_i}_{n=1}$ by ~\cref{eq:lp1};
		\ENDFOR
		\ENDFOR
		\RETURN $\{\mu_1^{(m)},\mu_i^{(m)}, \cdots, \mu_{N_k}^{(m)}\}$ and $\{{\mathbf{C}}_i^{n,(m)}\}^{z_i^{(m)}}_{n=1}$;
		\ENDFOR
		\RETURN global model $\Theta^{k}$ and global prototype set $\{\widetilde{\mathbf{C}}^{n}\}^z_{n=1}$;
	\end{algorithmic}
\end{algorithm}
Algorithm \ref{algorithm: AllModel} describes the specific optimization process of the multi-stage local model in one round of global optimization.
Lines 3 to 8 summarize the details of central server aggregation parameters during one round of local training.
Lines 9 to 18 describe the optimization process of the model for each local client, where $N_k$ clients execute the local model training in parallel.
For $M$ continuous learning tasks existing in the local clients, $M$ iterations are performed.
During one round of global optimization, the time complexity of the proposed method is determined by the number of incremental tasks and the number of local optimization rounds, which is about $\mathcal{O}(M(t+s))$.
For space complexity, GLDP only needs a small amount of space to save local prototypes and global prototypes, and does not need to save the training data related to the task of the previous stage.
%-------------------------------------------------------------------------
\section{Experiments}

We focus on the following two parts:
1) The effectiveness of GLDP on the \emph{spatial heterogeneity}.
2) The effectiveness of GLDP on the \emph{\textbf{spatio-temporal}} heterogeneity.

\subsection{Implementation Details}

\begin{table*}[!ht]
	\centering
	\caption{The $\mathcal{A}^{loc}$ (\%) of different FL methods on all datasets with different IFs. Note that we highlight the \textbf{best} results in bold and the \underline{second best} results in underlining.}
	\renewcommand{\arraystretch}{0.92}
	\scalebox{0.55}{
		\begin{tabular*}{25cm}{@{\extracolsep{\fill}}cccc|ccc|ccc|ccc|c}
			\toprule
			& \multicolumn{6}{c}{CIFAR10}& \multicolumn{6}{c}{CIFAR100}& \multicolumn{1}{c}{TinyImage}\\
			\cmidrule{2-7}\cmidrule{8-13}\cmidrule{14-14}
			& \multicolumn{3}{c}{[100, 4, 1]}&\multicolumn{3}{c}{[100, 5, 1]}&\multicolumn{3}{c}{[100, 20, 1]}&\multicolumn{3}{c}{[100, 30, 1]}& \multicolumn{1}{c}{[100, 20, 1]}\\
			\cmidrule{1-7}\cmidrule{8-13}\cmidrule{14-14}
			IF & 10 &50 & 100 & 10 &50 & 100 & 10 &50 & 100 & 10 &50 & 100 & 50\\
			\midrule
			FedAvg~\cite{mcmahan2017communication} &43.10&50.53&55.02&43.73&53.80&59.61&20.37&26.63&29.11&20.18&26.24&30.09&15.03\\
			FedRep~\cite{collins2021exploiting} &42.25&49.00&53.39&43.48&53.31&57.28&21.31&26.85&29.56&21.00&26.65&29.59&15.10 \\
			FedProx~\cite{li2020federated} &44.55&53.15&57.17&44.63&55.58&60.40&20.40&26.91&28.59&20.80&27.07&29.76&14.76  \\
			APFL~\cite{deng2020adaptive} &30.37&46.73&49.78&39.96&51.23&54.19&18.17&23.22&25.33&18.13&23.07&26.72&10.27  \\
%			FEDIC~\cite{shang2022fedic} &34.97&36.43&36.15 &33.78 &35.61  &36.52 &31.07&32.52&33.33&31.13&32.11& 32.59& \\
%			CReFF~\cite{2022Federated} &47.56 & 60.79&  65.04 & 47.25 &63.94 &68.93& 28.41& & &26.79 & 34.97&39.53&  \\
			\midrule
			GLDP-GP &\textbf{63.39}&\textbf{75.79}&\textbf{78.87}&\textbf{60.61}&\textbf{74.75}&\textbf{78.42}&\underline{35.61}&\underline{45.91}&\underline{48.56}&\textbf{34.70}&\underline{44.23}&\underline{47.68}&\underline{32.88} \\
			GLDP-LP &\underline{60.52}&\underline{75.20}&\underline{78.09}&\underline{56.98}&\underline{73.82}&\underline{77.25}&\textbf{37.05}&\textbf{49.14}&\textbf{52.79}&\underline{34.56}&\textbf{45.18}&\textbf{48.94}&\textbf{41.01}\\
			\bottomrule
	\end{tabular*}}
	\vspace{-0.5cm}
	\label{tab:example}
\end{table*}

\textbf{Datasets.}
We selected three popular benchmark datasets: CIFAR10, CIFAR100~\cite{krizhevsky2009learning}, and Tiny-ImageNet (TinyImage). When verifying the long-tailed distribution, we constructed the original balanced data into long-tailed distributions with IF = 10, 50, and 100 according to~\cite{cao2019learning}.
Similar to previous study~\cite{collins2021exploiting}, to construct a client-side spatially heterogeneous data partition, each client is assigned the same number of training samples but with different class distributions.

\textbf{Baselines.} We compare global-local dynamic prototype model (GLDP) model with FedAvg~\cite{mcmahan2017communication}, FedRep~\cite{collins2021exploiting}, FedProx~\cite{li2020federated}, APFL~\cite{deng2020adaptive}, FEDIC~\cite{shang2022fedic} and CReFF~\cite{2022Federated}. The first four personalized FL methods are used to solve the non-iid problem, and the last two methods alleviate the heterogeneous problem with long-tailed distribution.

\textbf{Hyperparameter.}
We employ ResNet18~\cite{He2016Deep} as the basic backbone for all datatsets. 
For all datasets, we set 100 clients when $M = 1$.
Each time the server-side randomly selects 10 clients to participate in training aggregation parameters. We set $M = 5$ for each client and a total of 20 clients.
For all methods, we set the number of local training epoch $t+s = 30$, the global training round $K = 50$.
The batch size is set to 32 for the CIFAR dataset and set 128 for TinyImage.
The weight decay is $1 e-4$ for local training.
The learning rate is initialized to be 0.01 and $\lambda$ is set to 0.5.
If we set 100 clients, each client contains 4 classes and 1 stage task data, recorded as [100, 4, 1].
We set $\beta$ = 0.5 for old classes in order to tradeoff the new and old sample knowledge of the same class and we set $\lambda$ = 0.5 to trade off the global and local prototype learning in Spatio-Temporal Heterogeneous Federated Learning (STHFL).

\textbf{Evaluation Metrics.}
We report the standard metric of method quality: accuracy, which is defined as the number of correctly predicted samples divided by the total number of samples.
To separately evaluate the performance of the local model and the global model, and the stability of the local model per client, we define as follows:\\
1) $\mathcal{A}^{glo}$ measure is the global model to perform verification tests on all client data samples.
\begin{equation}
	\begin{split}
		\mathcal{A}^{glo}=\frac{1}{N}\sum_{i=1}^{N}\frac{\sum\left(\operatorname{argMax}\left(\mathcal{F}_\Theta\left(X_{i}^{Test}\right)\right)==Y_{i}^{Test}\right)}{\left|\mathcal{D}_i^{Test}\right|}.
		\nonumber
	\end{split}
\end{equation}
2) $\mathcal{A}^{loc}$ measure is the local personalization model to perform validation tests on all client data samples.
\begin{equation}
	\begin{split}
		\mathcal{A}^{loc}=\frac{1}{N}\sum_{i=1}^{N}\frac{\sum\left(\operatorname{argMax}\left(\mathcal{F}_{\omega_i}\left(X_{i}^{Test}\right)\right)==Y_{i}^{Test}\right)}{\left|\mathcal{D}_i^{Test}\right|}.
		\nonumber
	\end{split}
\end{equation}
3) $\mathcal{A}_{i}^{sel}$ measure is the local personalization model to test the partial client participants.
\begin{equation}
	\begin{split}
		\mathcal{A}_{i}^{sel}=\frac{\sum\left(\operatorname{argMax}\left(\mathcal{F}_{\omega_{i}^{(m)}}\left(\hat{X}_{i}^{Test,(m)}\right)\right)==Y_{i}^{(m)}\right)}{\left|D_i^{(m)}\right|},
		\nonumber
	\end{split}
\end{equation}
where $D_i^{(m)} = \hat{D} _{i}^{Test,(m)}\bigcup D_{i}^{Test,(m)}$ and $\hat{D} _{i}^{Test,(m)}$ represent the set of samples from all previous stages.
Note that the FEDIC and CReFF method has no personalized model, thus $\mathcal{A}_{i}^{sel}$ use global model and no $\mathcal{A}^{loc}$ results.

\subsection{Spatial Heterogeneous FL Setting}

We conduct a set of experiments to verify the effectiveness of GLDP in the spatially heterogeneous FL and comparison with other advanced FL methods.
The GLDP-GP is the prediction obtained using the global prototype when validated using~\cref{eq:test}, while the result predicted using the local prototype is recorded as GLDP-LP.

\begin{figure}[!htb]
	\begin{center}
		\includegraphics[width=1.1\linewidth]{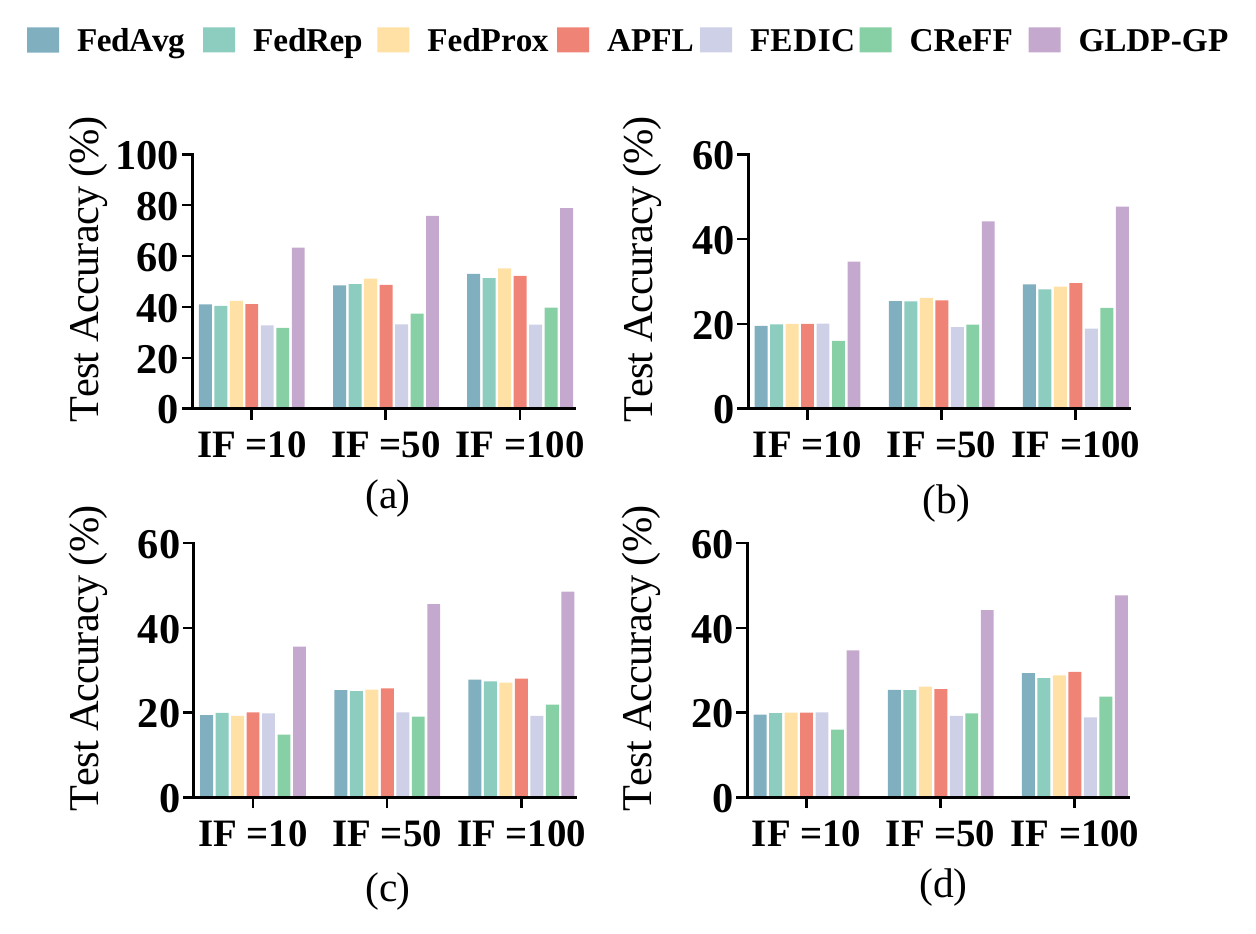}
	\end{center}
	\vspace{-0.7cm}
	\caption{The $\mathcal{A}^{glo}$ (\%) of different FL methods on CIFAR10 and CIFAR100 with different IFs. (a) [100, 4, 1]; (b) [100, 5, 1]; (c) [100, 20, 1]; (d) [100, 30, 1].}
	\label{fig:globalall}
	\vspace{-0.3cm}
\end{figure}

\noindent\textbf{Performance of \emph{Global} model.} 
\Cref{fig:globalall} depicts the comparison of the effect of GLDP with other federated learning methods addressing spatial heterogeneous.
The [N, S, M] denotes that the number of clients is $N$ and each client contains $S$ classes, and $M$ represents the stage tasks of each client.
We can observer that:
1) The GLDP can effectively alleviate the problem of \emph{spatial heterogeneity}, and the effect is significantly better than other personalized fine-tuning FL methods.
Since other methods just focus on personalized model fine-tuning to address data heterogeneity, the effect of the global model is ignored.
The GLDP benefits from the fusion of global and local prototype knowledge to trade off global-local models.
2) With the increase of \emph{spatial heterogeneity}, GLDP is relatively robust.
As the classes of each client increase, so do the heterogeneity.
The $\mathcal{A}^{glo}$ result of [100, 4, 1] in CIFAR10 is $78.87$, which is slightly different from the $78.42$ of [100, 5, 1].

\noindent\textbf{Performance of personalized \emph{Local} models.} 
\Cref{tab:example} lists the results of the local personalized model on all client-side sample data.
We can see that the proposed method outperforms the other FL studies.
From the results of verification using two prototypes, testing with local prototypes is superior to global prototypes because the former is auxiliary knowledge belonging to each client.
It can be seen that individualized knowledge of each client can mitigate inter-domain concept drift.
%Therefore, the results obtained when all clients are tested with personalized local models and local prototypes are outstanding.
%the test accuracy of all FL methods on the CIFAR100 dataset is lower than that of CIFAR10 due to the increased number of classes.
%Especially when validating the GLDP method using local prototype testing is better than the global prototype, because the former is an auxiliary knowledge that belongs to each client.

\subsection{Spatio-Temporal Heterogeneous FL Setting}

The incremental experiment setup is only training the current stage task data but testing samples of all previous and the current stages.

\noindent\textbf{Performance of \emph{Global} and \emph{Local} models.}
~\Cref{tab:glstage,tab:local1} record the results of GLDP compared with different FL methods when $M = 5$.
The presented results illustrate that our model performs the best for both global and local models after learning through several incremental tasks.
It also reflects that GLDP can train the stability global-local models.
Meanwhile, the performance of GLDP-GP is better than GLDP-LP
the performance of verification with local specific class prototypes to each client is better.

\begin{table}[!ht]
	\centering
	\vspace{-0.3cm}
	\caption{The $\mathcal{A}^{glo}$ (\%) of different FL methods on all datasets with different IFs. }
	\renewcommand{\arraystretch}{0.92}
	\scalebox{0.8}{
		\begin{tabular*}{16cm}{@{\extracolsep{\fill}}c c c| c c |c }
			\toprule
			& \multicolumn{2}{c}{CIFAR10}& \multicolumn{2}{c}{CIFAR100}& \multicolumn{1}{c}{TinyImage}\\
			\cmidrule{2-3}\cmidrule{4-5}\cmidrule{6-6}
			& \multicolumn{2}{c}{[20, 4, 5]}& \multicolumn{2}{c}{[20, 20, 5]}& \multicolumn{1}{c}{[20, 20, 5]}\\
			\cmidrule{1-6}
			IF & 50  & 100 & 50  & 100& 50\\
			\midrule
			FedAvg~\cite{mcmahan2017communication} &59.94&64.33&33.48&37.51&31.78 \\
			FedRep~\cite{collins2021exploiting} &62.68&66.91&33.94&37.63&32.14   \\
			FedProx~\cite{li2020federated} &59.23&62.60&32.50&36.12&31.82   \\
			APFL~\cite{deng2020adaptive} &60.32& 65.11&33.70&38.05&31.69   \\
			FEDIC~\cite{shang2022fedic} &34.34&33.43&21.76&22.41&29.99  \\
			CReFF~\cite{2022Federated} &45.12&44.32&30.69&35.09&27.54 \\
			\midrule
			GLDP-GP &\underline{65.22}& \underline{71.19}&\underline{37.35}&\underline{41.03}&\textbf{37.64} \\
			GLDP-LP &\textbf{65.36} & \textbf{71.32}&\textbf{37.64}&\textbf{41.57}&\underline{36.20} \\
			\bottomrule
	\end{tabular*}}
	\label{tab:glstage}
	\vspace{-0.1cm}
\end{table}

\begin{table*}[!ht]
	\centering
	\vspace{-0.6cm}
	\caption{The $\mathcal{A}^{loc}$ (\%) of different FL methods on all datasets with different IFs.}
	\renewcommand{\arraystretch}{0.92}
	\scalebox{0.7}{
		\begin{tabular*}{18cm}{@{\extracolsep{\fill}}cccccccccc}
			\toprule
			& \multicolumn{4}{c}{CIFAR10}& \multicolumn{4}{c}{CIFAR100}& \multicolumn{1}{c}{TinyImage}\\
			\cmidrule{2-5}\cmidrule{6-9}\cmidrule{10-10}
			& \multicolumn{2}{c}{[20, 4, 5]}&\multicolumn{2}{c}{[20, 5, 5]}&\multicolumn{2}{c}{[20, 20, 5]}&\multicolumn{2}{c}{[20, 30, 5]}& \multicolumn{1}{c}{[20, 20, 5]}\\
			\cmidrule{1-5}\cmidrule{6-9}\cmidrule{10-10}
			IF &50 & 100 &50 & 100 &50 & 100 &50 & 100 & 50\\
			\midrule
			FedAvg~\cite{mcmahan2017communication}  &38.77&39.26  &40.68 &41.49  &21.22&23.43&20.35&22.99&22.57   \\
			FedRep~\cite{collins2021exploiting} &60.97&65.27  &63.75 &63.04  &36.26&39.60&35.18&39.18&\underline{36.96}  \\
			FedProx~\cite{li2020federated} &38.80&40.52 &41.17 &41.11  &20.05&22.52&20.00&21.20&22.69  \\
			APFL~\cite{deng2020adaptive}    &36.02&37.94 &38.67 &40.73  &16.43&18.50&16.44&18.45&22.89  \\
			\midrule
			GLDP-GP &\underline{65.22}&\underline{71.19}&\underline{67.58}&\underline{65.02}&\underline{37.35}&\underline{41.03}&\underline{35.87}&\underline{41.07}&\textbf{37.64} \\
			GLDP-LP &\textbf{65.36}&\textbf{71.32}   &\textbf{68.10} &\textbf{65.61}&\textbf{37.64}&\textbf{41.57}&\textbf{36.21}&\textbf{41.43}&36.20\\
			\bottomrule
	\end{tabular*}}
	\label{tab:local1}
	\vspace{-0.4cm}
\end{table*}

\begin{figure*}[t]
	\begin{center}
		\includegraphics[width=0.9\linewidth]{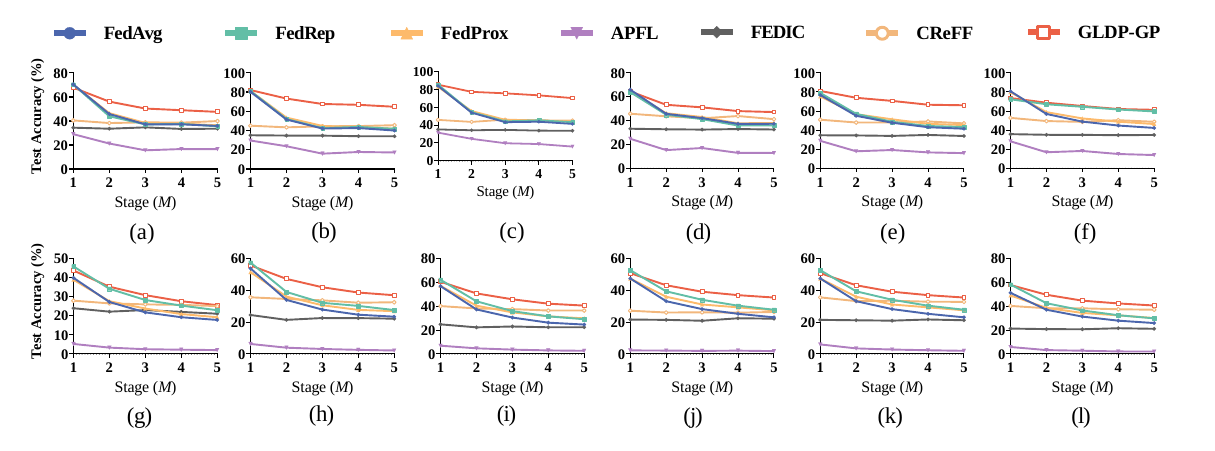}
	\end{center}
	\vspace{-0.7cm}
	\caption{Analysis of different stage tasks at client participants on CIFAR10 and CIFAR100 with $\mathcal{A}^{sel}$. 
		\textbf{CIFAR10:} IF = 10: (a) [20, 4, 5]; (b) [20, 5, 5]; IF = 50: (c) [20, 4, 5]; (d) [20, 5, 5]; IF = 100: (e) [20, 4, 5]; (f) [20, 5, 5];
		\textbf{CIFAR100:} IF = 10: (g) [20, 20, 5]; (h) [20, 30, 5]; IF = 50: (i) [20, 20, 5]; (j) [20, 30, 5]; IF = 100: (k) [20, 20, 5]; (l) [20, 30, 5];}
	\label{fig:stage}
	\vspace{-0.6cm}
\end{figure*}
%\begin{table*}[!ht]
%  \centering
%  \caption{The $\mathcal{A}^{loc}$ (\%) of different FL methods on all datasets with different IFs.}
%  \begin{tabular*}{18.2cm}{@{\extracolsep{\fill}}cccc|ccc|ccc|ccc|c}
	%    \toprule
	%     & \multicolumn{6}{c}{CIFAR10}& \multicolumn{6}{c}{CIFAR100}& \multicolumn{1}{c}{TinyImage}\\
	%     \cmidrule{2-7}\cmidrule{8-13}\cmidrule{14-14}
	%     & \multicolumn{3}{c}{[100, 4, 1]}&\multicolumn{3}{c}{[100, 5, 1]}&\multicolumn{3}{c}{[100, 20, 1]}&\multicolumn{3}{c}{[100, 30, 1]}& \multicolumn{1}{c}{[100, 20, 1]}\\
	%    \cmidrule{1-7}\cmidrule{8-13}\cmidrule{14-14}
	%    IF & 10 &50 & 100 & 10 &50 & 100 & 10 &50 & 100 & 10 &50 & 100 & 50\\
	%    \midrule
	%    FedAvg &32.52&38.77&39.26 & 33.51 & 40.68& 41.49 &21.22&23.43&&&\\
	%    FedRep &50.82&60.97&65.27 &51.44 &63.75 &63.04&36.26&39.60&&&  \\
	%    FedProx &33.37&38.80&40.52 &33.08 &41.17 &41.11&20.05&22.52&&&  \\
	%    APFL &30.37&36.02&37.94&29.98 &38.67 &40.73& 16.43&18.50&&&  \\
	%    %FEDIC & --& -- & --& --& -- \\
	%    %CReFF & -- &--  & -- &-- & -- \\
	%    \midrule
	%    GLDP &48.08&\underline{65.22}&\underline{71.19}&48.42&67.58&65.02&\underline{37.35}&\underline{41.03}&&& \\
	%    GLDP-LP &47.17&\textbf{65.36}&\textbf{71.32}&48.83&68.1&65.51&\textbf{37.64}&\textbf{41.57}& &&\\
	%    \bottomrule
	%  \end{tabular*}
%  \label{tab:local1}
%\end{table*}

\noindent\textbf{Analysis of incremental tasks for participating clients.}
\Cref{fig:stage} depicts the test accuracy of our selected participating clients in the final 10 global rounds.
The GLDP achieves the best accuracy on all settings, which validates that GLDP is effective in various data heterogeneity scenarios.
It can be observed that our method drops the slowest and the final result outperforms other competitive methods. 
Especially in the case of \Cref{fig:stage} (b), GLDP-GP is higher than the second best method 19.12\%.
According to the downtrend of these curves and the final results, it can be observed that our model performs better than other competing FL methods.
It demonstrates that GLDP can effectively adapt to the STHFL setting, enabling multiple clients to learn new data using prototype knowledge while mitigating the local model forgetting.
This also reflects that our model has a significant effect on solving the data distribution deviation from the pre-stage and post-stage within the client.

\begin{table}[!ht]
	\centering
	\caption{Impact of each component in GLDP. The experiments are conducted on CIFAR10 and CIFAR100 with IF = 50.}
	\renewcommand{\arraystretch}{0.95}
	\scalebox{0.8}{
		\begin{tabular*}{10cm}{@{\extracolsep{\fill}}c c c c c }
			\toprule
			& \multicolumn{2}{c}{CIFAR10}& \multicolumn{2}{c}{CIFAR100}\\
			\cmidrule{2-5}
			& \multicolumn{2}{c}{[20, 4, 5]}& \multicolumn{2}{c}{[20, 20, 5]}\\
			\cmidrule{1-5}
			Ablation & $\mathcal{A}^{glo}$ & $\mathcal{A}^{sel}$ &$\mathcal{A}^{glo}$ & $\mathcal{A}^{sel}$\\
			\midrule
			$\mathcal{L}_{\mathrm{Total}} - \mathcal{L}_{\mathrm{LP}} (\lambda = 0)$ &65.09 &63.47 &37.63 &30.08\\
			$\mathcal{L}_{\mathrm{Total}} - \mathcal{L}_{\mathrm{GP}}(\lambda = 1)$ &61.56 &60.52 &34.19 & 27.24\\
			$\mathcal{L}_{\mathrm{Total}} - \mathcal{L}_{\mathrm{GP}}-\mathcal{L}_{\mathrm{LP}}$ &62.68 &41.78 &33.94 & 27.50\\
			$\mathcal{L}_{\mathrm{Total}}$ &\textbf{65.22} &\textbf{ 64.61} & \textbf{37.64} & \textbf{36.82}\\
			\bottomrule
	\end{tabular*}}
	\vspace{-0.4cm}
	\label{tab:Necessity}
\end{table}

\noindent\textbf{Necessity of each component in GLDP.}
\noindent\Cref{tab:Necessity} displays the ${A}^{glo}$ and ${A}^{sle}$ results of GLDP after discarding some losses.
When $\lambda$ = 0, it means that the client only uses the global-local relationship loss, which is suitable for spatial heterogeneity learning.
When $\lambda$ = 1, it means that there is no interactive learning of prototype knowledge among clients, only local prototype learning in the pre-stage and post-stage.
The prediction results obtained by choosing different $\lambda$ are different.
We can see that removing any module leads to worse and unstable performance.
Especially in CIFAR100, $\mathcal{A}^{sel}$ performance drops 9.58\% without global-local prototype relation loss.
In addition, their joint absence can cause a further decrease on accuracy.
%Therefore, the $\lambda$ hyperparameter can be set according to the specific analysis requirements of the actual situation.
%It can be seen from the results that each module is critical to the performance of the GLDP model.

\section{Conclusion and Limitation}
In this paper, we propose a practical \emph{\textbf{spatio-temporal}} heterogeneity FL setting and develop a personalized global-local dynamic prototype fusion framework.
Specially, one-dimensional vector prototypes and low-dimensional representations of transmitted data are selected for aggregation to achieve local client-side data protection.
We address the domain drift problems of spatio-temporal heterogeneous setting and build the global-local models, thus attaining stronger generalization and stability.
Extensive experiments on representative benchmark datasets demonstrate the effectiveness of GLDP.

As in most FL simulation experimental setups, we set each client device to contain the same number of tasks.
However,  the number of tasks of different clients is different in real scenarios. 
For example, some large-scale hospitals have gradually increased case data and more tasks than small-scale hospitals.
In future work, we will set different numbers of tasks according to different types of client devices to suit the actual situation.
\bibliographystyle{elsarticle-num}
\bibliography{egbib}

%
%\begin{thebibliography}{00}
%
%%% \bibitem[Author(year)]{label}
%%% Text of bibliographic item
%
%\bibitem[ ()]{}
%
%\end{thebibliography}
\end{document}